\documentclass{article}

\usepackage{microtype}
\usepackage{graphicx}
\usepackage{subfigure}
\usepackage{booktabs} 

\usepackage{hyperref}

\usepackage[accepted]{icml2024}

\usepackage{amsmath}
\usepackage{amssymb}
\usepackage{mathtools}
\usepackage{amsthm}

\usepackage[capitalize,noabbrev]{cleveref}

\usepackage{enumitem} 

\theoremstyle{plain}
\newtheorem{theorem}{Theorem}[section]

\theoremstyle{definition}
\newtheorem{definition}[theorem]{Definition}

\theoremstyle{remark}

\usepackage[textsize=tiny]{todonotes}

\icmltitlerunning{A Framework for Inner Interpretability}

\begin{document}

\twocolumn[
\icmltitle{\textit{Position:} An Inner Interpretability Framework for AI \\ Inspired by Lessons from Cognitive Neuroscience}



\icmlsetsymbol{equal}{*}

\begin{icmlauthorlist}
\icmlauthor{Martina G. Vilas}{esi,goethe}
\icmlauthor{Federico Adolfi}{esi}
\icmlauthor{David Poeppel}{esi,nyu}
\icmlauthor{Gemma Roig}{goethe,hess}
\end{icmlauthorlist}

\icmlaffiliation{esi}{Ernst Strüngmann Institute for Neuroscience, Frankfurt am Main, Germany}
\icmlaffiliation{goethe}{Department of Computer Science, Goethe University, Frankfurt am Main, Germany}
\icmlaffiliation{nyu}{Department of Psychology, New York University, New York, USA}
\icmlaffiliation{hess}{The Hessian Center for AI, Hessen, Germany}

\icmlcorrespondingauthor{Martina G. Vilas}{martina.vilas@esi-frankfurt.de}

\icmlkeywords{Machine Learning, ICML}

\vskip 0.3in
]



\printAffiliationsAndNotice{}  

\begin{abstract}
Inner Interpretability is a promising emerging field tasked with uncovering the inner mechanisms of AI systems, though how to develop these mechanistic theories is still much debated.
Moreover, recent critiques raise issues that question its usefulness to advance the broader goals of AI.
However, it has been overlooked that these issues resemble those that have been grappled with in another field: Cognitive Neuroscience.
Here we draw the relevant connections and highlight lessons that can be transferred productively between fields.
Based on these, we propose a general conceptual framework and give concrete methodological strategies for building mechanistic explanations in AI inner interpretability research.
With this conceptual framework, Inner Interpretability can fend off critiques and position itself on a productive path to explain AI systems.
\end{abstract}

\section{Introduction}
\label{intro}
Inner Interpretability is an emerging subfield of Artificial Intelligence (AI) that is drawing increasing attention as models get larger and better.
Interpreting the internal mechanisms of these models in human-understandable terms is of great interest for scientific, engineering, and safety reasons \citep{rauker2023toward}.
However, despite interesting recent results, the field seems to be missing a conceptual framework that guides the development and analysis of these mechanistic explanations.
Although proposals have been made on how to quantify the alignment of human-intelligible high-level theories with the internal operations of neural networks \citep[e.g. causal abstraction approaches, see][]{geiger2023causal}, a more general framework for developing, discussing, analyzing, and refining these high-level mechanistic explanations is still needed.
The lack of such a conceptual framework has made this subfield vulnerable to critiques that question its usefulness to advance the broader goals of AI.

In addition, responses to these criticisms have not taken notice of how similar the issues raised are those to another field that has extensively dealt with them:
Cognitive Neuroscience \citep[but see][for more general links]{lindsay_testing_2023}.
In this position paper, we address these gaps.
\textbf{We explain the connections between issues in AI Inner Interpretability and those in Cognitive Neuroscience, and we propose that we can take advantage of these links to derive lessons and concrete conceptual and methodological strategies to interpret AI systems mechanistically}.

\underline{Overview:} 
The remaining sections are organized as follows.
Section \ref{ii_research} gives an overview of the field of Inner Interpretability and current critiques directed at it.
Section \ref{debates_cog_neuro} describes the issues that give rise to these critiques, draws parallels with longstanding issues in Cognitive Neuroscience, and explains how these have been tackled.
Section \ref{lessons_learned} draws on these lessons, proposes a conceptual framework for Inner Interpretability, and derives concrete methodological strategies.
Finally, in Section \ref{sec:implications} we explain how adopting this framework allows us to better integrate past and future studies and address critiques.

\section{AI inner interpretability research}
\label{ii_research}

Recent years have seen an increase in 
research aiming to understand the internal structural components, operations, and representations of deep neural networks.
This line of work has recently been given the name of \textit{inner interpretability} \citep[for a review see][]{rauker2023toward}.
A significant amount of work in this field is concerned with explaining how the inner mechanisms of these models give rise to their capabilities\footnote{In this paper, the term \textit{inner interpretability} is used to refer to this subset of work.}.
This differs from other lines of interpretability research where the behavior of the model is attributed to specific properties of the input or the training dataset \citep{bommasani2021opportunities}.

For example, inner interpretability studies have tried to characterize the model components \citep[e.g.][]{nelson2021mathematical, geva2020transformer, mcdougall2023copy, olsson2022context}, functions \citep[e.g.][]{merullo2023language, todd2023function}, and algorithms \citep[e.g.][]{zhong2023clock}, behind a variety of emergent behaviors.
Other work has developed methods to automate the discovery and analysis of activation sub-spaces \citep[e.g.][]{burns2022discovering},
circuits \citep[e.g.][]{conmy2023towards, lepori2023uncovering},
and internal representations \citep[e.g.][]{belrose2023eliciting, hernandez2023remedi}
that have a causal effect on the output of the model.

Inner interpretability work is motivated, on the one hand, by goals related to safety and transparency in AI \citep{casper2024blackbox}.
Mechanistic explanations may, for example, improve the predictability of model behavior
\citep{bommasani2021opportunities},
and enable editing out harmful or incorrect representations and steer decisions \citep[e.g.][]{hernandez2023remedi}.
On the other hand, the field could help improve aspects of model performance.
For instance, mechanistic discoveries can lead to efficiency improvements \citep{zhang2023instilling}, or the development of better architectures \citep[e.g.][]{akyürek2024incontext}.

Recent critiques of the field, however, paint a more pessimistic picture.
It has been questioned whether current methodological strategies, which are not yet well understood, will lead to any meaningful insights.
Critics believe that research practices in use are likely to lead to a false sense of understanding, which results in misleading or contradictory claims \citep[viz.][]{rauker2023toward}.
It has also been argued that these procedures can only achieve weak generalization to real-world problems or models \citep{doshi2017towards, rauker2023toward}.
More importantly, there seems to be a lack of clarity regarding the overarching questions of the field \citep{krishnan2020against}, and what it means to mechanistically understand a model.
Consequently, there is a heightened risk of letting the choice of research questions be guided by the availability of technology and heuristics \citep[i.e., a hardware-software lottery;][]{hooker2021hardware}.
These issues, if not addressed, put the field at risk of stagnation.

The outlined issues seem to arise from a lack of consensus on how to build mechanistic explanations, and how to evaluate and compare them through a common conceptual framework.
The field of Inner Interpretability is therefore in need of an analysis of its research practices and a general framework to guide them.
In this work, we propose that these tools can be adapted from the Cognitive Neuroscience field.
The rapid pace of current AI research often causes a disconnect from knowledge gained in adjacent fields.
In this case, Cognitive Neuroscience has confronted many of the issues that give rise to the critiques of Inner Interpretability, and has developed conceptual frameworks and methodological strategies to deal with them. 
In the next section, we set the groundwork to transfer these lessons to AI Inner Interpretability.

\begin{figure}[H]
\begin{center}
    \centering
    \includegraphics[scale=0.31]{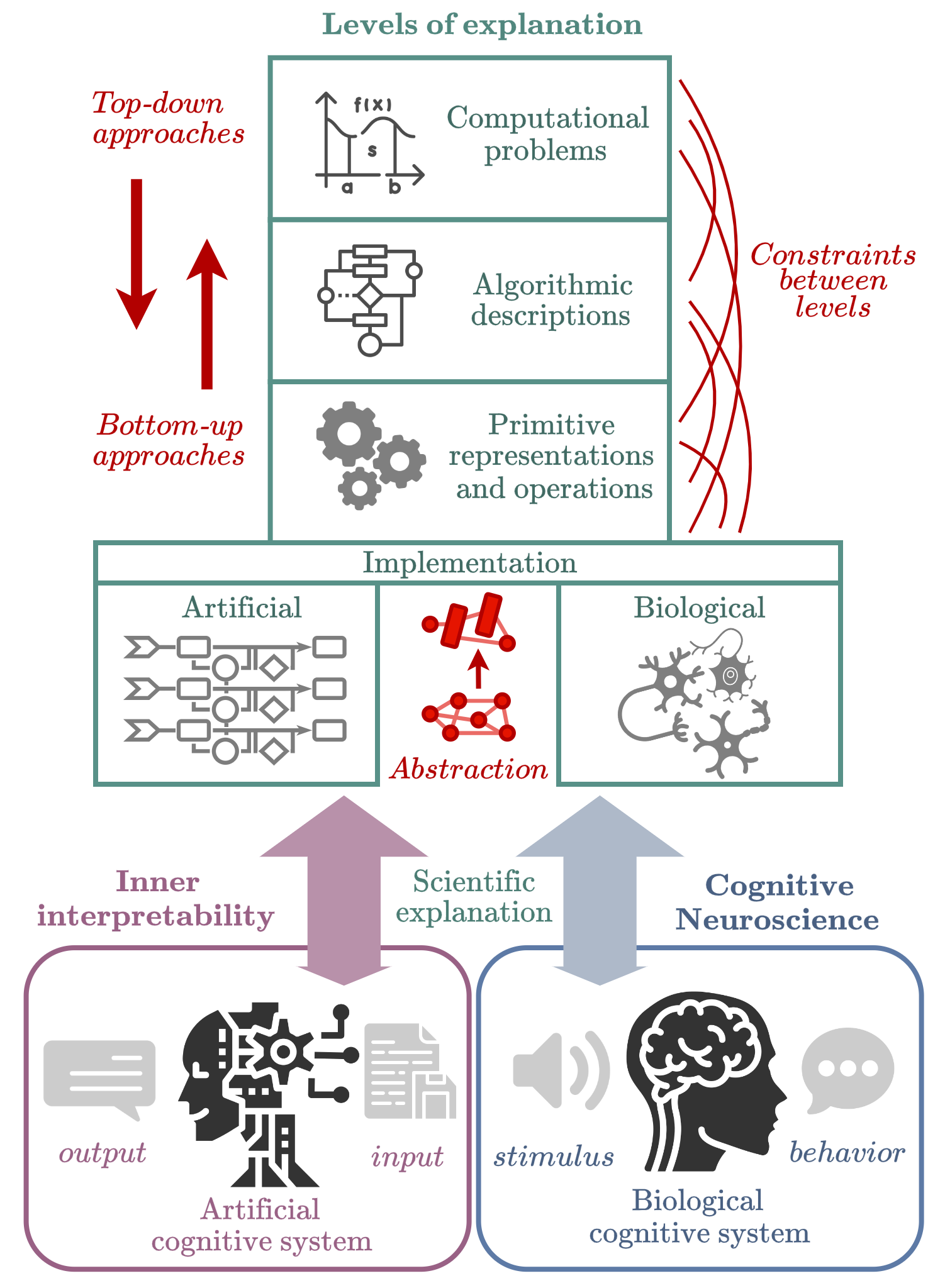}
\end{center}
\caption{The fields of Inner Interpretability and Cognitive Neuroscience aim to mechanistically explain the behavior of artificial and biological systems, respectively.
The multilevel explanatory framework proposed here draws out the parallels and suggests strategies that can be transferred between fields to tackle current issues in Inner Interpretability (shown in red).
}
\label{fig:interp-cogneuro}
\end{figure}

\vspace{-0.15in}

\section{Parallel issues between fields}
\label{debates_cog_neuro}

Cognitive Neuroscience and AI Inner Interpretability share similar goals, concepts, and methodology.
Both fields aim to uncover the mechanisms giving rise to the behavior of complex systems.
In both cases, researchers want to determine what capacities these systems possess, how they are implemented, and how they can be described in a human-understandable way.
The difference between the two fields, in principle, lies at the implementation level:  Cognitive Neuroscience must grapple with opaque biological substrates (e.g., human brains), whereas Inner Interpretability deals with the more accessible virtual substrate of artificial learning systems (e.g., transformer networks).

Naturally, obstacles for explaining these systems, and ways to overcome them, will be similar between fields.
Indeed, here we will demonstrate that overlapping issues exist.
Luckily, Cognitive Neuroscience is an older discipline than Inner Interpretability.
Therefore, the community has had time to confront these obstacles by developing conceptual frameworks and methodological strategies. 
This allows us to look at these proposals and apply insights to Inner Interpretability, to tackle the issues undermining the field's role in advancing the broader goals of AI.

In this section, we give some examples of these overlapping issues, namely, issues with mechanistic explanations, levels of abstraction, and bottom-up versus top-down approaches.
For each issue, we (i) show how it arises in Inner Interpretability and fuels critiques of the field, and (ii) explain how it played out and was tackled in Cognitive Neuroscience using multilevel conceptual frameworks \citep{marr1976understanding} and other methodological strategies that guide the development of mechanistic explanations.
Section \ref{lessons_learned} discusses how to apply these lessons from Cognitive Neuroscience to inner interpretability work.

\subsection{Issues with mechanistic explanation}
\label{mech_explain}

\textbf{Incomplete mechanisms in Inner Interpretability}.
Inner Interpretability is primarily concerned with uncovering the inner mechanisms of AI systems.
A classic notion of mechanistic explanation is an account of the relevant entities, activities, and organizational features (spatial and temporal relationships) that interact to have a causal effect in producing the phenomenon to be explained \citep{machamer2000thinking}.
In practice, however, a great number of inner interpretability studies seem to equate mechanistic explanations solely with localizing model components that have a causal effect on behaviors of interest. 
For example, researchers develop and make use of intervention methods such as ablation or activation patching, to find model components, such as neurons, sub-modules, and circuits, that have a causal link with the behavior \citep[e.g.][]{meng2022locating, vig2020causal}.
Yet, mechanistic proposals in these studies contain gaps and are thus \textit{incomplete}.
For instance, a neuron with a causal effect on the behavior may be localized, but its operation is left unspecified.
Or a causally relevant circuit may be mapped, but it is not decomposed into a sequence of entities performing activities \citep[for good examples of such decomposition see][]{merullo2023circuit, wang2022interpretability}.

Proposing incomplete mechanisms as explanations can lead to a false sense of understanding \citep{craver2006mechanistic}, and can easily result in mischaracterizing elements in the mechanism.
For example, a recent study probing the mechanisms of factual recall has shown that localizing an entity that carries factual information in a GPT model (e.g., MLP middle layers) does not mean that intervening in the operations of this entity will lead to the intuitive outcome: effective editing of this factual information \citep{hase2023does}.
This could be the result of a post-hoc misspecification of the operation performed by the MLP layers, in the context of an incomplete mechanistic proposal that did not spell out step-by-step how the output was produced.

\textbf{Attempts at complete mechanisms in Cognitive Neuroscience}.
Similarly to the field of Inner Interpretability, a significant number of studies in Cognitive Neuroscience operate under a lenient definition of mechanistic explanation \citep{ross_causation_2024}.
The term is often used to refer to findings showing a causal relationship between a neural component and a cognitive phenomenon, without providing a process description of how the outcome is produced.
However, researchers have long argued that this type of work is insufficient to build mechanistic explanations 
\citep[viz.][]{krakauer2017neuroscience}.
For instance, previous studies have shown the difficulty of inferring cognitive function from neural recordings \citep{poldrack2006can}, and the insufficiency of uncovering necessary and sufficient neural circuits for building mechanistic explanations \citep{gomez2017causal}.

To build better and more complete mechanistic explanations, the strategy of multilevel analysis was introduced \citep[more on this in Section \ref{lessons_learned};][]{marr1976understanding}.
Multilevel explanations include comprehensive functional characterizations of the behavior (\textit{what} is the system doing and \textit{why}), algorithmic descriptions of how the function is computed (\textit{how} is it doing it), and decomposition of the algorithms in a list of fine-grained and human-interpretable primitive operations and representations implemented in the neural substrate (see Fig. \ref{fig:interp-cogneuro}).
A complete multilevel explanation results in a characterization of the relevant brain components, their activities and interactions, that implement the capacity of interest.

More recent work has pointed out that mutual constraints between levels provide an avenue to construct more complete mechanistic explanations in practice \citep{danks2013moving, love2021levels}.
That is, results at one level can suggest what to look for at other levels.
The higher levels provide a conceptual and formal structure that can guide the search for and characterization of neural mechanisms \citep{griffiths2010probabilistic, krakauer2017neuroscience}.
For example, the location of circuits in the brain that are causally involved in speech processing has been known for a long time, but it was not until theoretically and empirically motivated computational steps were hypothesized that a better understanding of the functions of each of the neural structures composing the circuit was achieved \citep{hickok2007cortical}.
In analogy to the \textit{factual recall} example discussed previously, if a theoretically motivated sequence of primitive representations and operations had been mapped to the uncovered circuit for factual recall, perhaps the role assigned to the MLP layers would have been more accurate and interventions would have produced the predicted result.

\subsection{Issues with levels of abstraction}
\label{abstraction}

\textbf{Weak motivation for abstractions in Inner Interpretability}.
Mechanisms can be explained at different levels of abstraction at the implementation level (e.g., neurons, circuits, modules, representational trajectories\footnote{Levels of abstraction are not to be equated with levels of physical organization (i.e. spatio-temporal scale), nor with the levels of explanation of the multilevel framework. These are orthogonal.}; see Fig. \ref{fig:interp-cogneuro}, bottom).
That is, complete explanations can be given at levels of abstraction that ignore various details at other levels.
For example, explanations of how large language models can retrieve facts have been given at the level of sub-module operations, without referring to the neurons, layers, and non-linearities composing them \citep{chughtaisumming, geva2023dissecting}. 
Importantly, some levels of abstraction may lead to explanations that are more human-intelligible and can be efficiently uncovered across model sizes.

In practice, work on inner interpretability typically chooses the level of abstraction not based on mechanistic principles but rather arbitrarily (e.g., based on previous studies), without examining their implications.
Weak motivations for abstractions may block progress in building a robust understanding of the behavior.
For example, some assume that human-interpretable representations are to be found at the level of neurons, based on previous empirical findings \citep[e.g.][]{hernandez2021natural}.
However, recent studies suggest that relevant features may be encoded in superposition \citep{elhage2022toy}, representing a switch in abstraction level to `directions in activation space'.
In addition, it has been suggested that finding these explanations at microscopic levels of abstraction may not be computationally feasible in large models \citep{adolfi2024complexity, zou2023representation}.

\textbf{Attempts at choosing better abstractions in Cognitive Neuroscience}.
Neurobiological processes can also be characterized at different levels of abstraction and the choice is consequential \citep{barack2021two}.
The problem of choosing an appropriate level of abstraction has been discussed in Cognitive Neuroscience in the context of a \textit{mapping problem}: to build mechanistic explanations, the basic parts of neurobiology (e.g., synapses, neurons, brain regions) must be mapped onto the basic operations and representations of human cognition \citep{poeppel2016maps}.
For this match to succeed, the right level of abstraction for the basic parts of the brain must be discovered.

In the context of multilevel explanations \citep[see Section \ref{mech_explain};][]{marr1976understanding}, 
the level of primitive operations and representations can be conceptualized as depicting the set of cognitive `parts', while the implementation level encodes the set of neurobiological components \citep[viz.][]{poeppel2016maps}.
The selection of primitive candidates is motivated by theories of cognitive functions that have been formally and empirically validated.
But their choice is also constrained by the type of operations that can be implemented in the brain.
The reverse is also true: the decomposition into neural components at the implementation level is constrained by the list of plausible primitives.
Therefore, the choice of abstraction both at the level of primitives and at the implementation level can be guided by the quality of the match it affords between neural and cognitive `parts'.
For instance, \textit{segmentation} operations were long hypothesized as computational primitives of speech recognition.
To understand their implementation in the brain, cognitive neuroscientists found it useful to abstract away from neural circuit connectivity to focus instead on the level of oscillations of neuronal ensembles \citep{giraud_cortical_2012,poeppel_speech_2020}.
This abstraction has led to productive research programs.

Comparably, in Inner Interpretability, some recent approaches have proposed to adjust the level of abstraction of high-level theories (e.g. of the variables in causal models) to better align with empirical data \citep{geiger2023causal}. 
More work is needed to also guide this abstraction at the implementation level.

\subsection{Issues with bottom-up versus top-down approaches}
\label{bu-td}

\textbf{Overoptimistic bottom-up approaches in Inner Interpretability}.
A frequent distinction in the field of Inner Interpretability is between \textit{top-down} and \textit{bottom-up} research methodologies.
The term \textit{top-down} is linked to work mapping pre-defined and human-interpretable representations and operations to model components \citep[e.g.][]{kim2018interpretability, meng2022locating, mu2020compositional}.
Recently, it has been shown that some findings in this line of work can be misleading when assumptions are not well-tested \citep[e.g.][]{bolukbasi2021interpretability}, since alternative mechanisms may lead to the same empirical observation.
As an alternative, a bottom-up approach named \textit{mechanistic interpretability}\footnote{The term \textit{mechanistic interpretability }may cause confusion, as both top-down and bottom-up research study mechanisms.}
was introduced to study models ``without a priori theories" \citep{interp_dreams}.
They propose to decompose the network into the smallest elements possible, thoroughly investigate their functions and interactions by observing, perturbing, and describing them, and work upward to build abstractions from these foundations until their role in the behaviors of the model can be explained \citep{interp_dreams, olsson2022context}.

However, bottom-up approaches are not free of assumptions and, if left unexamined, these can easily lead to roadblocks, false understanding, and non-generalizable claims.
On the one hand, a choice is being made regarding which neural component to analyze, at which level of abstraction, what conditions to perturb, which behaviors to analyze as responses, etc.
In addition, it heavily relies on the interpretations made by a human observer.
Automatic interpretability methods are not free of these assumptions either.
The neurons-versus-directions example discussed in the last section illustrates the assumed levels of abstraction in automatic discovery methods.
On the other hand, these methods carry the risk of making inferences that do not scale up to the capacities and models of interest that originally motivated the research.
Given the vast space of research strategies without a priori theories, bottom-up studies start by tackling small problems in small networks that are easily interpretable with available tools.
However, the ultimate goals of understanding are related to the more complex capacities of large AI systems.
Most work assumes, without guarantees, that these research strategies will eventually scale beyond toy problems and models.

\textbf{Methodology-aware approaches in Cognitive Neuroscience}.
Whether \textit{top-down} or \textit{bottom-up} approaches are more effective in discovering mechanistic explanations has been discussed in Cognitive Neuroscience since its origins (see appendix Fig. \ref{fig:s1}).
Top-down approaches start by defining and decomposing the behavior of interest computationally.
Algorithmic candidates for these computations can then be linked to neural processes \citep[e.g.,][]{krakauer2017neuroscience,egan2018function}.
Bottom-up approaches first thoroughly describe and manipulate neural parts and activities, and then try to infer the cognitive capacity they implement \citep[e.g.,][]{buzsaki_braincognitive_2020,churchlandComputationalBrain1999}.
Like in Inner Interpretability, bottom-up approaches were advanced in response to fears of being misled by inaccurate theories.

Radical bottom-up approaches have highlighted the benefits of carrying out comprehensive brain mapping efforts at the finest levels of detail \citep[i.e., building a `connectome';][]{sporns_human_2005}.
Such efforts (e.g., the Human Connectome Project) were expected to turn low-level descriptions of the brain into high-level explanations of cognitive abilities.
The idea was that looking for regularities at the level of neural `stuff' would eventually reveal their underlying mechanistic organization.
However, these approaches are said to have ``overpromised and underdelivered" \citep{gomez-marin_promisomics_2021}.
There are at least two reasons for this failure.
First, there is rarely a one-to-one mapping between neural parts (e.g. circuits) and the algorithmic or functional descriptions of the cognitive capacities they implement \citep[viz.][]{gunaratne_variations_2017}. 
Second, bottom-up approaches are not free of assumptions.
These are needed to narrow down vast search spaces that cannot be explored exhaustively. 
The selection of neural parts and processes for investigation is necessarily based on such preconceptions.
Candidate cognitive operations and representations are also selected for exploration based on explicit or implicit assumptions.
Lack of explicit assumptions does not mean that assumptions are not being made about these aspects.
Explicit assumptions can be examined and tested.
Implicit or unexamined ones can easily result in misleading conclusions.

Cognitive Neuroscience has learned this lesson the hard way.
Various radical top-down and bottom-up approaches were put forth \citep[e.g.,][]{buzsaki_braincognitive_2020,krakauer2017neuroscience,niv_primacy_2021} and many of the promises have been overstated on both sides \citep[viz.][]{gomez-marin_promisomics_2021}.
But it is now clear that the simultaneous execution of combined approaches is necessary to discover their invariants and reach useful mechanistic explanations \citep{poeppel2020against}.
This more pluralistic process takes advantage of the mutual constraints between implementation and other levels of explanation (see Fig. \ref{fig:interp-cogneuro}, right) to arrive at consistent mechanistic theories.
In addition, to reduce the impact of incorrect assumptions and theories, a variety of methodological strategies have been proposed to rigorously test the validity of these conjectures.
We will discuss how to apply these to the field of Inner Interpretability in the next section.

\section{A framework for Inner Interpretability}
\label{lessons_learned}

In this section, we describe a conceptual framework where the lessons from Cognitive Neuroscience discussed in the previous sections can be translated into methodological strategies for AI Inner Interpretability.
Our running example will be \textit{factual recall}, a frequently studied topic in Inner Interpretability \citep{hernandez2023remedi, meng2022locating, geva2023dissecting, chughtaisumming, yu2023characterizing}.
We will illustrate, without loss of generality, how the strategies we propose can be applied to study the inner mechanisms of language transformers that implement the capacity to recall facts.

\subsection{Building multilevel mechanistic explanations}

Previous work has already pointed out the usefulness of applying a multilevel conceptual framework \citep[viz.][]{marr1976understanding} for better analyzing and comparing the performance of machine learning models \citep{hamrick2020levels}.
Here, we show that a multilevel analysis of capacities also provides a useful conceptual structure to investigate their inner mechanisms.
Each level offers a qualitatively different description of the mechanism under study, and as such each level employs a specialized terminology and provides a different angle of analysis (see appendix \ref{separability} for a discussion on their separability). 
A productive research program in Inner Interpretability makes use of
mutual constraints across the levels to arrive at a complete mechanistic explanation.
Next, we explain how to locate each level of explanation in Inner Interpretability research projects, and how to use mutual constraints between levels to converge on useful mechanistic explanations.

\subsubsection{Computational Problems}
\label{comp_level}
To build a mechanistic explanation, it is necessary to define and thoroughly characterize the phenomena to be explained \citep{craver2006mechanistic}.
In the framework of multilevel explanation (see Fig. \ref{fig:interp-cogneuro}), a systematic behavior of interest (e.g., factual recall) is selected and described at the \textit{computational} level \citep{marr1976understanding}.
A computational description gives a functional specification of the capacity underlying the observed behavior (i.e., it describes \textit{what} the system is doing). 
Here, the capacity can be characterized as an information-processing task where the system maps inputs to outputs.
This level also formalizes the input and output domains, and may specify additional properties or parameters of the mapping \citep[viz.][]{shagrir_marrs_2017}.

The difference between observed behavior and underlying capacity is important.
It is intuitive to observe some complex-looking model behavior (e.g., the classification of images of different animals using an abstract category such as `animal') and infer an interesting capacity of the model (e.g., the ability to build rich representations that abstract away from particular animals such as cats or dogs).
However, the same behavior can be the consequence of different underlying capacities.
The propensity of AI models to exploit `shortcuts' means that often some of the true underlying capacities turn out to be uninteresting.
For instance, cats and dogs can be distinguished from inanimate objects by building abstract representations, but this may also be achieved by exploiting contextual cues given by confounds in training datasets.
Ignoring these issues regularly leads to claims that models possess interesting capacities, followed by more rigorous experimentation eliciting behaviors that evidence their absence \citep[viz.][]{mitchell_why_2021,mitchell_debate_2023,bowers_deep_2022}.

Consider the example of \textit{factual recall},
a capacity intuitively described as the recall of truthful knowledge about entities in the world.
For instance, a fact might be `Paris is the capital of France', and the retrieval of `Paris' in response to the prompt `The capital of France is [---]' is an example of factual recall behavior of the model.
This intuitive verbal definition is insufficient to carry out a scientific analysis of the mechanisms that enable its emergence in a large language model.
To be more precise, we can define factual recall as the capacity to recall an attribute (e.g., Paris) when prompted with a subject (e.g., France) and a relationship (e.g., capital), from a particular knowledge domain (e.g., political geography).
A formal definition of the capacity could be constructed as follows (to be refined iteratively).

\begin{definition}[Facts and fact domains]
    \label{def:fact}
    A \textit{fact} is a 3-tuple $F = (S, R, A) \in \mathcal{D}$ , where $S$ is a subject, $R$ is a relation, $A$ is an attribute, and $\mathcal{D} = \{F_1, F_2, ..., F_n\}$ represents a \textit{fact domain}. Incompleteness of a fact tuple is denoted with $\bot$ in the corresponding component.
\end{definition}

\begin{definition}[Factual recall] \label{def:prob-fact-recall}
\begin{itemize}[leftmargin=0pt]
    \setlength\itemsep{0.1pt}
    \item[]
    \item[] \textit{Computational problem:} $\mathcal{D}$\textsc{-FactualRecall}
    \item[] \textit{Input:} An incomplete \textit{fact} tuple (Def. \ref{def:fact}), $F_I = (S, R, \bot)$ corresponding to a complete fact $F = (S, R, A) \in \mathcal{D}$, where $\mathcal{D}$ is a constant fact domain.
    \item[] \textit{Output:} A completion $F_C$ of $F_I$ such that $F_C \in \mathcal{D}$.
\end{itemize}
\end{definition}

A computational level explanation should also specify \textit{why} the behavior occurs in its specific context \citep[][see also resource-rational analysis: \citealt{lieder2020resource}]{shagrir_marrs_2017}. 
That is, it should explain how environmental properties constrain and shape the function of the system.
In our example, a context where the model is expected to deliver truthful information (e.g., a chatbot interacting with users) encourages the emergence of factual recall.

Computational-level descriptions (see Fig. \ref{fig:interp-cogneuro}, top) can be tested both formally and empirically.
Descriptions at this level take the form of system capacities as computational problems (e.g., Def. \ref{def:prob-fact-recall}).
But, not all computational problems that can be written down describe capacities that are possible in practice (e.g., they are uncomputable or intractable; \citealp{wareham_systematic_1998}; or fall outside the expressive power of the AI architecture; \citealp[viz.][]{strobl_transformers_2023}).
Overlooking this can lead to explanations that are inconsistent between levels.
For instance, one could inadvertently propose a computational-level description of a capacity that is outside the class of problems that the model architecture (e.g., transformer) can solve \citep[viz.][]{strobl_average-hard_2023,strobl_transformers_2023}, yielding inconsistency between computational and implementation levels.
Indeed, the choice of the computational-level description can be guided by examining the functions that can be yielded by implementational properties of the model, such as its architectural design, optimization strategy, or initialization procedure.
In sum, proposals at the computational level can be formally tested to determine if they are possible in the real world \citep[e.g., using the formal tools of theoretical computer science;][]{garey1979computers,downey_fundamentals_2013,wareham_systematic_1998}, under relevant constraints such as the model architecture.

Empirically, researchers can study the behavior of interest to characterize the reliability and flexibility of the underlying capacity, and to determine any relevant restrictions. 
Behaviors of foundation models, for instance, cannot be taken for granted, even for simple problems \citep{bommasani2021opportunities}.
Established benchmarks, often named after capacities (``language understanding", ``commonsense reasoning"), do not always test these fully and might not address questions of interest \citep[viz.][]{mitchell_debate_2023}.
Customized benchmark datasets are needed to properly test computational-level proposals \citep[e.g.,][]{moskvichev_conceptarc_2023}.

\subsubsection{Algorithmic descriptions} 

Multilevel explanations contain a description of the algorithms and data structures that implement the computational theory \citep{marr1976understanding}.
At this level, a sequence of human-understandable steps that make the capacity possible is spelled out.
Algorithmic descriptions can also be provided in the form of causal models \citep{geiger2023causal}.
Algorithms cannot be ascertained from the study of the capacity alone because a single capacity can be realized by different algorithms \citep[e.g.,][]{zhong2023clock}.

The Attribute Extraction procedure (Algorithm \ref{alg:prop-extract}) is a candidate algorithm for the computational problem $\mathcal{D}$\textsc{-FactualRecall} (Def. \ref{def:prob-fact-recall}), inspired by \citet{chughtaisumming}.
It consists of a series of steps that extract attributes related to the subject and relationship entities included in a factual recall prompt, and outputs the attribute that is highly associated to both the subject and relationship.
The algorithm starts from an incomplete fact tuple containing input representations, subject $S$ and relation $R$.
The \textsc{AttributeBoost} subroutine takes as input an element of the incomplete tuple, and outputs a vector whose components correspond to a pre-defined vocabulary, where the entries associated with attributes of the input are numerically boosted.
The \textsc{AttributeCombine} step combines the boosted attribute vectors. 
\textsc{AttributeMax} then outputs the maximum-value attribute of the combined vectors. 
This description clarifies in human-understandable terms how the capacity of factual recall may be implemented step-by-step.

\begin{algorithm}[H]
   \caption{Factual Recall via Attribute Extraction}
   \label{alg:prop-extract}
\begin{algorithmic}
   \STATE {\bfseries Input:} incomplete fact tuple $(S, R, \bot) \in \mathcal{D}$
   \STATE $H_s = \text{AttributeBoost}(S)$
   \STATE $H_{s,r} = \text{AttributeBoost}(S, R)$
   \STATE $H_r = \text{AttributeBoost}(R)$
    \STATE $H_i = \text{AttributeCombine}(H_s, H_{s,r}, H_r)$
    \STATE $A = \text{AttributeMax}(H_i)$
    \STATE {\bfseries Return} attribute $A$
\end{algorithmic}
\end{algorithm}

\subsubsection{Primitive representations and operations}
The third level from the top specifies how the algorithm is executed using \textit{primitive operations and representations} \citep{marr1976understanding, poeppel2016maps}.
Primitives are the basic building blocks of the system, without which the phenomena cannot occur \citep{poeppel2016maps}. 
They must be grounded theoretically, supported by abundant empirical evidence, and be realizable at the implementation level.
Primitives, which can emerge through training (i.e., \textit{post hoc}), should not be confused with the basic components of model architectures that were programmed before learning (e.g., single neurons, activation functions).
It is possible that certain \textit{post-hoc} primitives can emerge consistently in distinct models as a result of similar inductive biases given, for example, by architectures, training datasets, or learning rules.

\begin{figure}[h]
\begin{center}
\centerline{\includegraphics[width=\columnwidth]{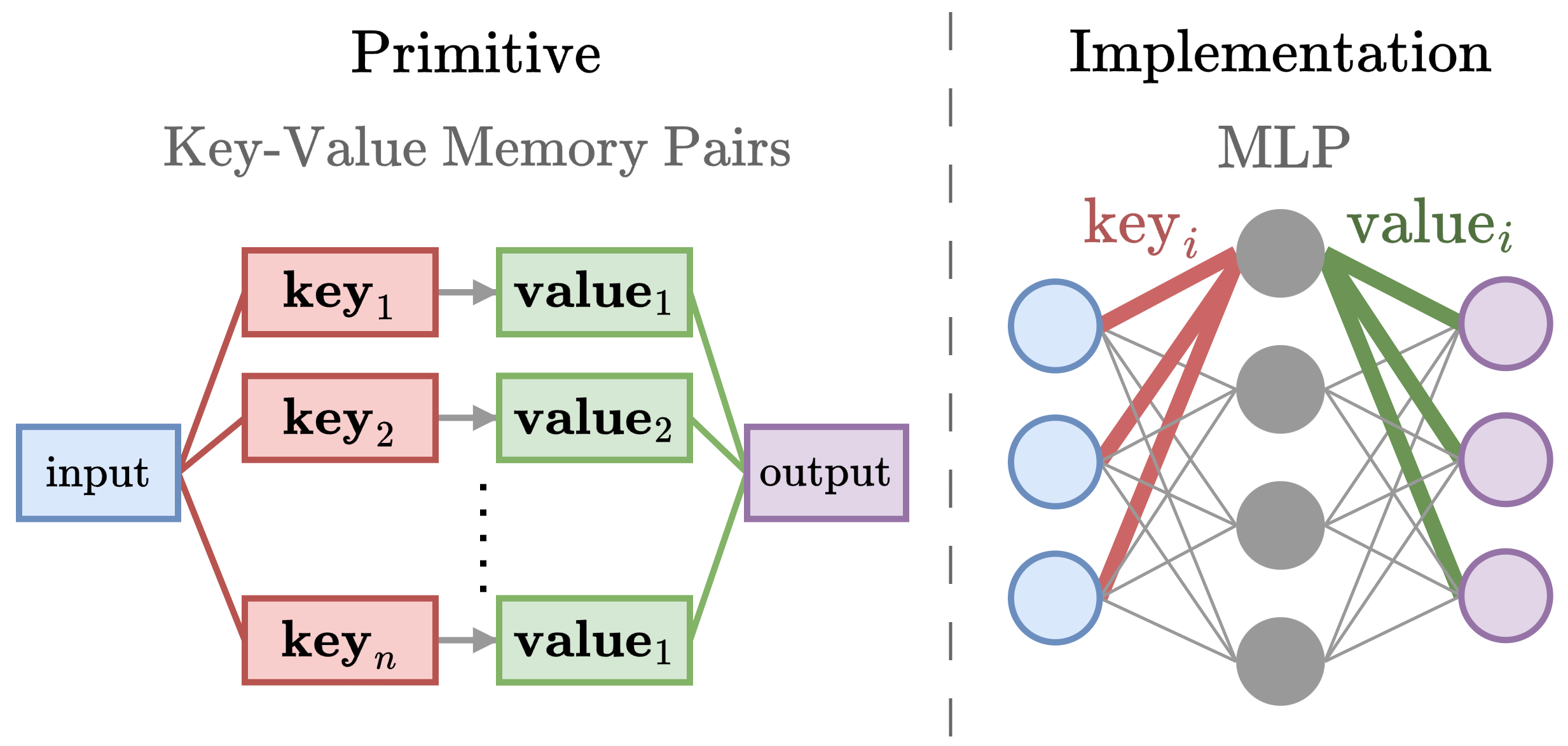}}
\end{center}
\vspace*{-0.2in}
\caption{Examples of the \textit{primitives} and \textit{implementation} levels using the key-value memory pairs system in MLP layers.
}
\label{fig:kvm}
\end{figure}

In the factual recall example, a primitive candidate for $\textsc{AttributeBoost}$ is a \textit{key-value memory pair system} (see Fig. \ref{fig:kvm}). 
Generally, these systems are comprised of a set of paired vectors called \textit{key} and \textit{value}, where the key of a pair detects a pattern in the input (e.g. subject, relation, or their combination) and the value outputs associated information (e.g., attributes).
Key-value systems emerge in transformer models across domains, models, tasks, and sub-modules \citep[e.g.,][]{geva2020transformer, meng2022locating, vilas2023analyzing_vit}.
Moreover, their existence is also theoretically motivated \citep{sukhbaatar2019augmenting}.
This makes the \textit{key-value memory pair system} a suitable candidate for an operational primitive.

\subsubsection{Implementation}
The \textit{implementation level} characterizes how the primitives are implemented in the model.
For this, the network must be decomposed at a certain level of abstraction.
These choices need not match those of the architecture design before training \citep[viz.][]{pylyshyn1978computational}.
The decomposition can be guided by the proposals at higher levels of explanation including the list of primitives.
Exploratory work and the use of localization heuristics can also be used as guidance. 
Identifying model parts that have a causal effect on the output may help constrain the possible primitive operations, by examining the computations that can be implemented by these components \citep{bechtel2007mental}.
For example, \textit{concept localization} efforts can be used to determine where relevant information emerges, and subsequently the function responsible for this emergence can be explored.
Upper levels can then be revised to accommodate these empirical discoveries.

In the context of factual recall example, studies have demonstrated that key-value memory pair systems can both theoretically and practically be implemented by MLP or self-attention layers \citep[viz.][]{dar-etal-2023-analyzing, geva2020transformer, vilas2023analyzing_vit}.
Therefore, these sub-modules are suitable candidates to describe the implementation of key-value system primitives (see Fig. \ref{fig:kvm}).

\subsection{Building hypotheses and conducting severe tests}

Previous work has already emphasized the need for rigorous hypothesis-testing procedures in interpretability research \citep{leavitt2020towards, rauker2023toward}.
While an increasing number of studies are adopting these practices, there is ample room for improving how hypotheses are derived and empirically tested when evaluating mechanistic proposals.
To begin with, candidate hypotheses should be tightly linked to mechanistic conjectures.
Continuing with the factual recall example, one possible hypothesis could be: ``extraction of subject attributes is conducted in subject tokens via key-value memory pair systems implemented by mid-layer MLP modules". 
This only tackles one element of the mechanism but in a realistic experiment, hypotheses should be made about all components of the proposal.

Concrete and falsifiable empirical predictions are derived from mechanistic hypotheses.
In the factual recall example, the following empirical predictions could be made: 
``one or more vectors in the first MLP parameter matrix will encode values that highly activate with subject tokens. The corresponding vectors of the second MLP matrix will produce distributional updates that promote attributes related to the subject."
Empirical predictions should differ from those of a reasonable baseline condition.
For instance, they should minimally differ from those made about untrained models.
Similarly, it is important to assess whether any other plausible mechanism could produce the same empirical predictions and to compare candidates.
Competing mechanistic proposals should have true potential for explaining the capacity (\citealp{wilson_ten_2019}; see e.g., \citealp{zhong2023clock}).

Hypotheses can be evaluated using \textit{severe tests} \citep{mayo2018statistical}, which have been recently introduced to the field of Cognitive Neuroscience to better evaluate how empirical practices and findings contribute to the assessment of models of human cognition \citep{aktuncc2014severe, bowers2023importance}.
Concretely, empirical observations and the methods used to evaluate them should have a high probability of falsifying a mechanistic proposal if it is incorrect.
Severe tests can be extended to the field of AI \citep{bowers2023importance}. 
Inner interpretability researchers should ask themselves: if the hypothesized mechanism is absent, how likely is the method to reveal its absence? 
Severe tests require work examining the adequacy of the methods themselves, for instance, by exploring whether they can falsify mechanistic hypotheses about a system whose design principles are known.

\subsection{Designing experiments}

Insights about the adequacy of mechanistic proposals can be gained by investigating input conditions that modulate the behavior of the model \citep{craver2006mechanistic}.
Certain environmental conditions can precipitate the occurrence of a behavior, inhibit it \citep[e.g., failure modes, see][]{hardcastle2015marr}, or modulate it.
Experiments can mimic these conditions and evaluate whether hypothesized mechanisms effectively account for the modulation of the model's capabilities.
Failure to provide adequate explanations and predictions suggests missing or erroneous elements in mechanistic proposals.
As an example, models may fail to retrieve a fact because it was not available in the training set and thus not learned, or because the prompt was constructed in a way that led to failures in outputting the information \citep{jiang2020can}.
Presumably, the mechanisms behind these failures are different,
and mechanistic proposals should be able to differentiate them.

In addition to exploring these controlled experimental conditions, mechanisms should also be able to explain behavior under \textit{naturalistic conditions}.
For example, during model deployment, the way users construct prompts to extract factual information is more variable than those of controlled experiments.
To be useful for the broader goals of AI, studies should ideally demonstrate that they can explain the manifestation of the capacity in these naturalistic settings.

\subsection{Testing invariances}

Multilevel explanations are often expected to generalize across conditions.
Some of these conditions are explicit in the computational description.
For instance, mechanisms are expected to remain invariant across input subdomains.
In the factual recall case, the capacity could in principle be investigated separately for various fact subdomains (e.g., geography, politics, literature).
However, the focus is to be kept on discovering the invariants, since these capacities are hypothesized to be manifestations of a more general capability of the system.

Other invariant conditions are left implicit in the computational explanation.
For example, which kind of models are expected to implement the capacity is often left unsaid.
Researchers may want to generalize their claims to narrower or broader classes of models (e.g. GPT-3, Large Language Models, Transformers). 
Typically, mechanisms invariant to initialization and hyperparameter values are sought \citep[these often affect the inner workings of models;][]{zhong2023clock}.
Similarly, researchers look for mechanisms that remain valid when the models are scaled up.
Whether mechanistic proposals can scale to describe larger models accurately can be supported or challenged by formal analyses \citep[viz.][]{arora_computational_2009}. 

\subsection{Refining mechanistic proposals and conceptual frameworks}

Mechanistic proposals are to be continuously improved with new theoretical and empirical findings.
When a change to one of the levels of explanation is made, all other levels need to be iteratively revised by imposing mutual constraints.
Importantly, mechanistic proposals do not have to be complete to be tested or to be useful more generally.
Mechanistic sketches (i.e. explanations that contain gaps), can point to productive avenues for research \citep{bechtel2007mental, craver2006mechanistic}, within and across research projects, \textit{as long as their role in more comprehensive explanations is on the horizon}.
Indeed, mechanistic proposals of factual recall in the Inner Interpretability literature have been increasingly refined and filled in, evidencing a productive line of work.

Moreover, it is essential to continuously revisit the utility of the adapted conceptual framework and methodological strategies for guiding the development of mechanistic explanations.
Newer reappraisals of the multilevel framework, and how it is adapted to Inner Interpretability research, may lead to the formulation of better mechanistic explanations. 
For example, it has recently been suggested that separate levels should be added to the multilevel framework to describe how intelligent systems learn, and how such learning is the product of evolution \citep{poggio2012levels}.
However, adapting these levels to inner interpretability research entails challenges, since the learning constraints/goals and the conceptualization of evolution processes in biological systems cannot be easily extrapolated to that of artificial systems.
Future work is needed to conceptualize this adaptation.

\section{Implications for Inner Interpretability}
\label{sec:implications}
The framework presented here can be used to examine the mechanisms of AI models regardless of the complexity of their architecture, training data, or task performed.
Studies probing more complex systems and capacities may especially benefit from this framework, as it promotes simplified and human-understandable explanations that abstract from irrelevant implementational details and are generalizable to a variety of controlled and naturalistic contexts.
More broadly, as detailed in the following subsections, the framework helps elucidate the state of the Inner Interpretability field and offers insights on how to move forward.
In addition, 
adopting this framework helps tackle common criticisms of the field.

\textbf{Situating previous studies.}
Studies in Inner Interpretability that appear hard to reconcile given their rationale and methods now have a coherent relationship in the multilevel framework proposed here.
For example, some work can be understood as providing algorithmic-level descriptions of model capacities, as well as how they are realized at the implementation level \citep[e.g.,][]{chughtaisumming, merullo2023circuit, wang2022interpretability, zhong2023clock}.
Similarly, approaches like causal abstraction and structural equation modeling have been developed to evaluate how algorithmic descriptions fit the empirical data acquired at the implementation level \citep{beckers2019abstracting, chalupka2017causal, geiger2023causal, rubenstein2017causal}.
Other work can be viewed as focusing on uncovering primitive operations and representations across domains, architectures, and sub-modules \citep[e.g.,][]{geva2020transformer, olsson2022context, vilas2023analyzing_vit}.
Still, another body of work can be framed as developing heuristics for localization of causally relevant components at the implementation level, via automated circuit- and feature-finding procedures \citep[e.g.,][]{burns2022discovering, conmy2023towards, gurnee2023findingneurons}.

\textbf{Identifying research gaps.} 
The framework also sheds light on potentially productive research avenues.
Future directions derive directly from the aspects of the framework that are absent in current and past studies.
Overall, there is a need for work formalizing capacities and theoretically analyzing their computational viability.
A better understanding of the assumptions of common inner interpretability methods is also needed.
More generally, studies proposing mechanistic sketches that span all four levels of the explanation are essential for progress.

The framework can also illustrate how a new research domain could be studied, or make it easy to locate gaps in some less-studied lines of research. 
For example, it can be used to identify aspects for improvement in studies probing the formation of abstract representations in neural networks.
As explained in section \ref{comp_level}, instead of providing descriptions of behaviors, future work could focus on formalizing capacities and later determining with theoretical and empirical work the behaviors that would be adequate reflections of them in a variety of scenarios. 
At the algorithmic level, Inner Interpretability studies have traditionally analyzed at which stages of the model hierarchy abstract representations emerge \citep[e.g.][]{ilin2017abstraction}. 
However, work investigating the sequence of algorithmic steps that build these abstract representations is still needed. 
Similarly, no work has been carried out to understand if the primitives put in use to form abstract representations differ from those of more concrete concepts, as suggested by research in Cognitive Neuroscience. 
At the implementation level, no studies have probed the right abstraction level or distributive nature of the model components supporting abstract representations.

\textbf{Addressing criticisms.}
This framework offers guidance on how to addresses critiques arguing that the field lacks consensus and clarity on what a mechanistic understanding of a system is and how to build it. 
Regarding the criticism that inner interpretability methods have limited applicability to practical problems or realistic models, this work shifts the focus to studying well-defined capacities that link to real-world applications, and provides concrete strategies to do so.
In addition, this framework helps avoid building a false sense of understanding and making misleading claims, by (i) encouraging a more comprehensive characterization of model capacities via multilevel analysis, inter-level constraints, and converging lines of evidence, (ii) emphasizing the use of methods involving severe tests to yield robust findings with assumption-aware interpretations.

\section*{Impact statement}
The framework can be used to research the safety of the internal mechanisms of AI models. 
For example, the computational level can facilitate the fine-grained formalization and evaluation of the set of capacities (not behaviors) that the system should, or should not, possess according to the safety standards. 
The algorithmic level encourages spelling out and investigating if the algorithms that the system deploys are acceptable and guarantee safe behaviors of the model. 
In turn, the primitives and implementation levels call for research on how fragile the uncovered mechanisms are, and how they can be edited for safety reasons.

Overall, effective inner interpretability techniques should ultimately be useful to make systems safer to interact with, and more energy-efficient.
However, other lines of work are more urgent than Inner Interpretability to address these issues.
Tackling training data curation problems \citep[viz.][]{birhane_into_2023} and fostering responsible use with respect to carbon emissions \citep[viz.][]{luccioni_counting_2023,luccioni_power_2023} are concrete avenues to deal with well-documented issues, whereas the benefits of inner interpretability in this context are currently more speculative.

\section*{Acknowledgment}
This project was partly funded by the Ernst Strüngmann Foundation and the German Research Foundation (DFG) - DFG Research Unit FOR 5368. 
We are grateful for access to the computing facilities of the Center for Scientific Computing at Goethe University, and of the Ernst Strüngmann Institute for Neuroscience.

\bibliography{main}
\bibliographystyle{icml2024}

\newpage
\appendix
\onecolumn

\section{Critical concepts and references from Cognitive Neuroscience} 
\label{concepts}

\begin{figure}[H]
\begin{center}
    \centering
    \includegraphics[scale=0.4]{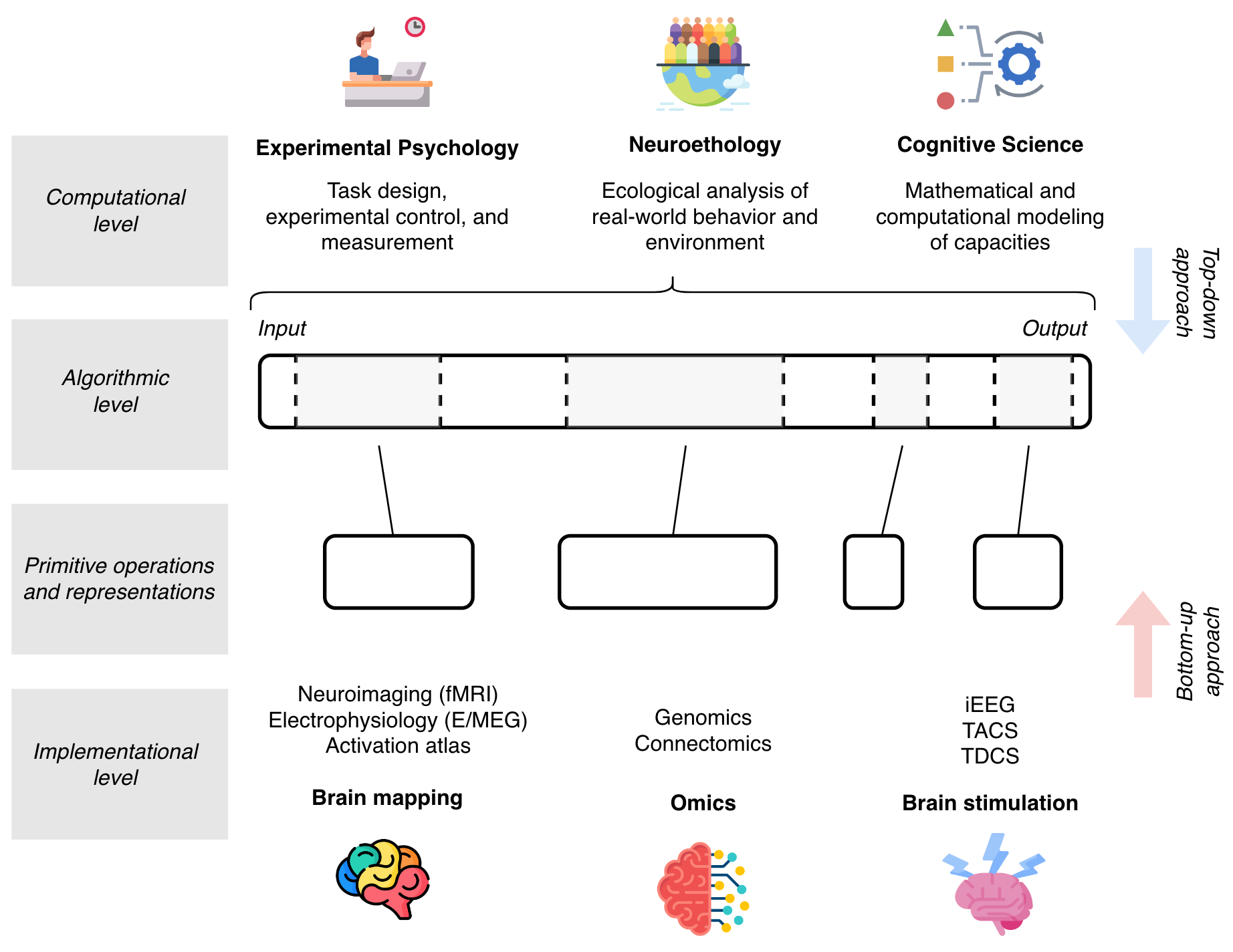}
\end{center}
\caption{Schematic of the key techniques deployed by different disciplines in Cognitive Neuroscience organized according to whether they promote top-down or bottom-up approaches to the discovery of mechanistic explanations. Note that only radical (top-down/bottom-up) approaches propose to reach inner levels through a one-directional use of these techniques. In practice, the discovery of mechanistic explanations involves a healthy combination of techniques from top-down and bottom-up approaches.
}
\label{fig:s1}
\end{figure}

Below we provide a discussion of concepts used in the Cognitive Neuroscience field that are critical for the proposed AI Inner Interpretability framework:

\begin{itemize}

    \item \textbf{Mechanistic explanations} in neuroscience are described as detailing the entities (or parts), activities (or operations), properties, organizational features (both temporal and spatial), and causal relationships among these components, that produce the target phenomena \citep{bechtel2007mental, craver2006mechanistic, kaplan2011explanatory, machamer2000thinking}.
    The entities or parts ``[...] are the things that engage in activities" \citep{machamer2000thinking}, and ``[...] are the structural components of the mechanism" \citep{bechtel2007mental}.
    The activities or operations ``[...] are the producers of change" \citep{machamer2000thinking}, and ``[...] refer to processes or changes involving the parts" \citep{bechtel2007mental}.
    
    \item \textbf{Mechanistic sketches} are incomplete models of a mechanism, and signal that further work is needed \citep{craver2006mechanistic, machamer2000thinking}.
    These sketches contain gaps where certain components of the mechanistic explanation, such as entities or activities, are missing.
    Gaps are sometimes masked by filler terms. For example, terms like cause, encode, produce, and represent ``[...] are often used to indicate a kind of activity in a mechanism without providing any detail about how that activity is carried out" \citep{craver2006mechanistic}.

    \item The target of explanation in Cognitive Neuroscience is the neural implementation of a cognitive capacity. 
    A \textbf{capacity} is an underlying ability of a system or organism to transform certain input to output states \citep{egan2018function}.
    It can be fully defined at the computational level of analysis by specifying the input domain and the function that maps inputs to outputs.
    A \textbf{behavior} can be understood as the concrete and observable manifestation of a capacity.
    In the literature, the term behavior is often used to refer to a behavioral phenomenon, a set of concrete actions by a system or organism that may reflect the performance of an underlying capacity. 
    Defining what constitutes a behavior is challenging \citep{calhoun2021behavior}.
    Behaviors of interest are chosen based on observations and theoretical arguments that they represent a manifestation of the target cognitive capacity.
    
    \item It has been argued that cognition can be mechanistically explained as a sequence of \textbf{computational operations} performed over \textbf{representations} \citep{bechtel2007mental}.
    Although the definition and properties of the term `representation` continue to be a matter of debate in the field, they are frequently conceptualized as states of the neural system that carry information about external objects or events relevant to the capacity being explained \citep{bechtel2007mental}.
    A list of the elementary mental representations and operations has been called the \textit{human cognome} and corresponds to the \textbf{primitive} units on analysis of the cognitive sciences ``[...] without which they could not account for the elementary phenomena of their field" \citep{poeppel2016maps}. 

    \item Different \textbf{levels of explanation} can be used to analyze the mechanisms of intelligent systems. \citet{marr1976understanding} propose that four ``nearly independent" levels of description can be used to study machines that solve an information processing problem: ``(1) that at which the nature of a computation is expressed; (2) that at which the algorithms that implement a computation are characterized; (3) that at which an algorithm is committed to particular mechanisms; and (4) that at which the mechanisms are realized in hardware".
    \citet{marr1976understanding} put special emphasis on the importance of the computational level, which is often neglected in Cognitive Neuroscience studies.
    Later work removed the third level from the multilevel framework, while in this work we re-conceptualize it as the encoding of cognitive parts that need to be mapped to other levels.
    Furthermore, recent reappraisals of this framework have highlighted the importance of utilizing mutual constraints among the levels to achieve better explanations of the cognitive system, a point we support in this position paper.

    \item Neuroscience studies can choose to provide a mechanistic explanation at different \textbf{levels of neural organization}.
    For example, some studies seek to explain cognition ``[...] as the result of operations on signals performed at nodes in a network and passed between them that are implemented by specific neurons and their connections in circuits in the brain" \citep{barack2021two}, while others explain cognition ``[...] as the result of transformations between or movement within representational spaces that are implemented by neural populations" \citep{barack2021two}.
    The levels of neural organization are orthogonal to the levels of explanation.
    Although only at the implementation level the neural components are explicitly examined, their choice implies commitments that constrain the possible explanations at other levels.

    \item In Cognitive Neuroscience, it has been greatly debated whether neural mechanisms should be studied before or after having decomposed and analyzed the behavior of interest at the computational and algorithmic levels (i.e. \textbf{bottom-up} or \textbf{top-down} approaches, see Fig. \ref{fig:s1} and section \ref{bu-td}).
    For example, in favor of a top-down approach, \citet{krakauer2017neuroscience} argue that ``higher-level concepts are needed to understand neuronal results" (higher-level concepts are those derived from behavioral work) and provide a variety of examples of how ``behaviorally driven neuroscience yields more complete insights". 
    In contrast, bottom-up proponents like \citet{buzsaki_braincognitive_2020} argue that ``[...] most of our behavior-related terms emerged before and independent of neuroscience, and there is little guarantee that these terms correspond to circumscribed brain mechanisms".
    In their view, the field should instead ``[...] start with the brain (independent variable) and define descriptors of behavior (dependent variables) that are free from philosophical connotations" \citep{buzsaki_braincognitive_2020}.
    In short, we can ``[...] recast the inherent tension between these epistemic procedures as that between \textit{What is a mechanism for X?} versus \textit{What is Y a mechanism for?}" \citep{poeppel2020against}.
    Recently, it has been argued that more pluralistic approaches that combine methods from both research positions may lead to more robust mechanistic theories: ``[...] we might view the process of doing research in the field of cognitive neuroscience as the iterative abduction of certain kinds of mechanistic theories about human capacities" \citep{poeppel2020against}, where abduction ``[...] jointly captures the process by which a set of candidate explanations is generated from observations and background knowledge (sometimes termed abduction proper), and how the choice among them is justified" \citep{poeppel2020against}.

\end{itemize}

\section{Key ideas and models of bottom-up and top-down approaches in Cognitive Neuroscience}

\section{Separability of the levels of explanation}
\label{separability}

Certain levels of the multi-level explanation framework are more easily distinguishable from neighboring levels than others. 
The computational and algorithmic levels provide descriptions that are formally distinguished in Computer Science, and as such cannot be easily confounded. 
In contrast, the algorithmic and primitive levels use a similar vocabulary. Determining whether a particular sub-computation is a primitive relies on the theoretical and empirical evidence supporting its role as a building block of the system. Finally, the implementation level is also easily distinguishable from other levels, as it provides descriptions using the terminology used when designing the model (e.g. MLP layers, self-attention heads, activation functions, etc.).

\section{Using AI to understand biological cognitive functions}

Previous work has also leveraged the similarities between AI research and Cognitive Neuroscience by employing models and techniques from the AI field to better understand biological cognitive functions.
Among other proposals, it has been suggested that deep learning models can serve as tools for testing cognitive theories \citep{storrs2019deep}.
For example, they can be used to probe the learning rules, goals, and anatomical properties of the brain \citep{richards2019deep}.
Moreover, they can be employed to test the structure and content of cognitive representations in the human brain \citep{sucholutsky2023getting}.
Regarding AI techniques, it has recently been proposed that tools to interpret neural networks can be used to test the methods in neuroscience \citep{lindsay_testing_2023}.


\end{document}